\documentclass[letterpaper]{article} 
\usepackage{aaai2026}  
\usepackage{times}  
\usepackage{helvet}  
\usepackage{courier}  
\usepackage[hyphens]{url}  
\usepackage{graphicx} 
\usepackage{natbib}  
\usepackage{caption} 
\usepackage{algorithm}
\usepackage{algorithmic}
\usepackage{amsmath} 
\usepackage{booktabs}
\usepackage{amsfonts}
\usepackage{multirow}

\title{Enhancing Robustness of Offline Reinforcement Learning Under Data \\ Corruption via Sharpness-Aware Minimization (Student Abstract)}

\author{
    Le Xu\textsuperscript{\rm 1}\thanks{Accepted as an Oral Presentation at the AAAI 2026 Student Abstract and Poster Program (SAPP).},
    Jiayu Chen\textsuperscript{\rm 2}
}
\affiliations {
    \textsuperscript{\rm 1}Tsinghua University,\\
    \textsuperscript{\rm 2}The University of Hong Kong\\
    xu-l22@mails.tsinghua.edu.cn, jiayuc@hku.hk
}

\begin{document}

\maketitle

\begin{abstract}
Offline reinforcement learning (RL) is vulnerable to real-world data corruption, with even robust algorithms failing under challenging observation and mixture corruptions. We posit this failure stems from data corruption creating sharp minima in the loss landscape, leading to poor generalization. To address this, we are the first to apply Sharpness-Aware Minimization (SAM) as a general-purpose, plug-and-play optimizer for offline RL. SAM seeks flatter minima, guiding models to more robust parameter regions. We integrate SAM into strong baselines for data corruption: IQL, a top-performing offline RL algorithm in this setting, and RIQL, an algorithm designed specifically for data-corruption robustness. We evaluate them on D4RL benchmarks with both random and adversarial corruption. Our SAM-enhanced methods consistently and significantly outperform the original baselines. Visualizations of the reward surface confirm that SAM finds smoother solutions, providing strong evidence for its effectiveness in improving the robustness of offline RL agents.
\end{abstract}

\section{Introduction}
Offline reinforcement learning (RL) learns policies from static datasets \cite{levine2020offlinerl}, a vital paradigm for real-world applications where online interaction is infeasible. We frame this problem within a Markov Decision Process (MDP) $(\mathcal{S}, \mathcal{A}, P, r, \gamma)$, where the goal is to learn a policy $\pi(a|s)$ from a static, and potentially corrupted, dataset $\mathcal{D}$. A major challenge is the prevalence of data corruption, which can severely degrade performance.

Our work builds upon Implicit Q-Learning (IQL) \cite{iql}, a powerful offline algorithm that learns a Q-function and a value function $V$ via expectile regression, and then extracts a policy through advantage-weighted behavioral cloning. Due to this design, IQL has shown notable resilience to some forms of data corruption. However, even IQL and its robust variant, RIQL \cite{riql24}, exhibit significant performance drops under observation and mixture corruptions.

We hypothesize this vulnerability arises because data corruption creates sharp, unreliable minima in the loss landscape. To this end, we are the first to apply Sharpness-Aware Minimization (SAM) \cite{pf2020} as a general-purpose optimizer for offline RL. SAM seeks flat minima, which are known to improve generalization and robustness. By integrating SAM, we guide training towards more stable solutions without altering the core logic of the base algorithm. Our experiments demonstrate that this approach consistently enhances the performance of IQL and RIQL on challenging D4RL benchmarks under both random and adversarial corruption, with reward surface visualizations confirming that SAM finds smoother, more robust solutions.

\section{Methodology}
Our core proposal is to integrate Sharpness-Aware Minimization (SAM) as a plug-and-play optimization module for offline RL agents. This approach is motivated by the hypothesis that data corruption induces sharp minima in the value function's loss landscape. Models converging to these sharp minima tend to be less robust, as small perturbations in the input data can lead to large errors in value estimation, ultimately degrading policy performance. Instead of altering the algorithm's loss functions, we replace the standard optimizer (e.g., Adam) with a SAM wrapper to explicitly seek out flatter, more generalizable solution regions.

\paragraph{SAM Optimization Process.}
The SAM optimizer seeks parameters $\theta$ that lie in neighborhoods with uniformly low loss. It achieves this via a two-step minimax procedure for a given loss function $L(\theta)$:
\begin{enumerate}
    \item \textbf{First step (ascent):} It computes the gradient $\nabla_{\theta} L(\theta)$ to find an adversarial weight perturbation $\hat{\epsilon}(\theta)$ that locally maximizes the loss, effectively probing the sharpness of the landscape. It then ascends to the perturbed weights $\theta' = \theta + \hat{\epsilon}(\theta)$.
    \begin{equation}
        \hat{\epsilon}(\theta) = \rho \frac{\nabla_{\theta} L(\theta)}{\|\nabla_{\theta} L(\theta)\|_2 }
    \end{equation}
    \item \textbf{Second step (descent):} It computes the gradient $\nabla_{\theta} L(\theta')$ at this "worst-case" perturbed point and uses this gradient to update the original parameters $\theta$.
\end{enumerate}
This process penalizes sharpness by minimizing the highest loss value within the local neighborhood, forcing the optimizer to find flat regions where the loss remains low even after perturbation.

\paragraph{Integration with Value Function Learning.}
We implement SAM as a custom PyTorch `Optimizer` class that wraps a base optimizer like Adam, allowing for seamless integration. Based on ablation studies (see Appendix), we found that applying SAM exclusively to the value function network yields the most significant and stable improvements. Therefore, in our main experiments, the standard update step for the value function's parameters is replaced by SAM's two-step procedure, ensuring that the learned value function is robust to perturbations in its parameter space.

\section{Experiments}
We conduct experiments to answer: Can applying SAM to SOTA offline RL algorithms improve their robustness against observation and mixture corruption?

\paragraph{Experimental Setup.} We evaluate on three continuous control environments from the D4RL benchmark \cite{fu2021d4rldatasetsdeepdatadriven}: `halfcheetah-v2`, `walker2d-v2`, and `hopper-v2`, using the `medium-replay` datasets for each. We test on two challenging corruption types: \textbf{Observation} and \textbf{Mixture}, under both \textbf{random} and \textbf{adversarial} settings. Our base algorithms are IQL and RIQL. All corruptions are applied to 30\% of the dataset. Detailed corruption methods are described in the Appendix. All experiments are conducted over three random seeds, and we report the mean and standard deviation of the normalized scores.\cite{fu2021d4rldatasetsdeepdatadriven}

\paragraph{Main Results.}
Tables \ref{tab:random_corruption_styled} and \ref{tab:adversarial_corruption_styled} show our main results. Under both random and adversarial corruption, applying SAM consistently improves the performance of both IQL and RIQL across all environments and corruption settings. The average performance gains are significant, demonstrating that SAM is an effective technique for enhancing robustness. The improvements are particularly pronounced in the more challenging Mixture corruption setting, validating SAM's ability to handle complex noise distributions.

\begin{table}[ht]
\centering
\resizebox{\columnwidth}{!}{%
\begin{tabular}{@{}ll cc cc@{}}
\toprule
\multirow{2}{*}{Environment} & \multirow{2}{*}{Corruption} & \multicolumn{2}{c}{IQL} & \multicolumn{2}{c}{RIQL} \\
\cmidrule(lr){3-4} \cmidrule(lr){5-6}
& & Naïve & SAM (Ours) & Naïve & SAM (Ours) \\
\midrule
Halfcheetah & Observation & 21.01(3.01) & \textbf{33.33(1.49)} & 26.03(3.61) & \textbf{33.74(2.48)} \\
            & Mixture     & 20.93(4.21) & \textbf{33.02(2.81)} & 22.08(2.47) & \textbf{32.06(2.17)} \\
\midrule
Walker2d    & Observation & 24.74(7.10) & \textbf{31.75(8.87)} & 30.48(11.82) & \textbf{30.93(9.56)} \\
            & Mixture     & 25.90(6.25) & \textbf{31.68(2.93)} & \textbf{26.92(10.90)} & 19.55(17.04) \\
\midrule
Hopper      & Observation & 58.42(1.49) & \textbf{73.21(11.72)} & 44.09(8.35) & \textbf{53.19(2.01)} \\
            & Mixture     & 55.86(18.37)& \textbf{63.42(12.29)} & 54.20(13.80) & \textbf{67.36(10.61)} \\
\midrule
\multicolumn{2}{l}{Average score $\uparrow$} & 34.47 & \textbf{44.40} & 33.97 & \textbf{39.47} \\
\bottomrule
\end{tabular}%
}
\caption{Average Performance under random corruption}
\label{tab:random_corruption_styled}
\end{table}

\begin{figure}[ht]
    \centering
    \includegraphics[width=0.48\columnwidth]{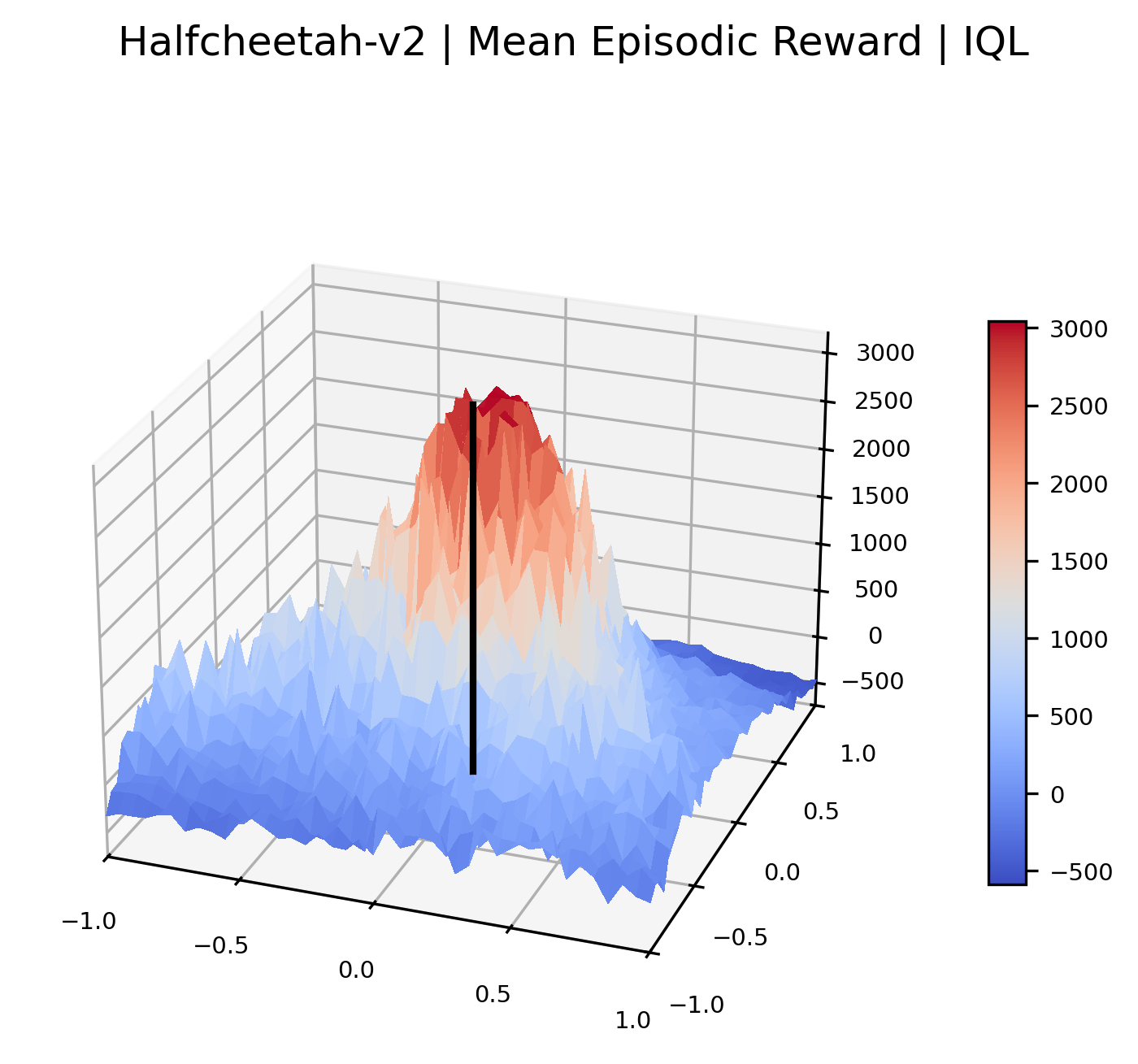}\hfill
    \includegraphics[width=0.48\columnwidth]{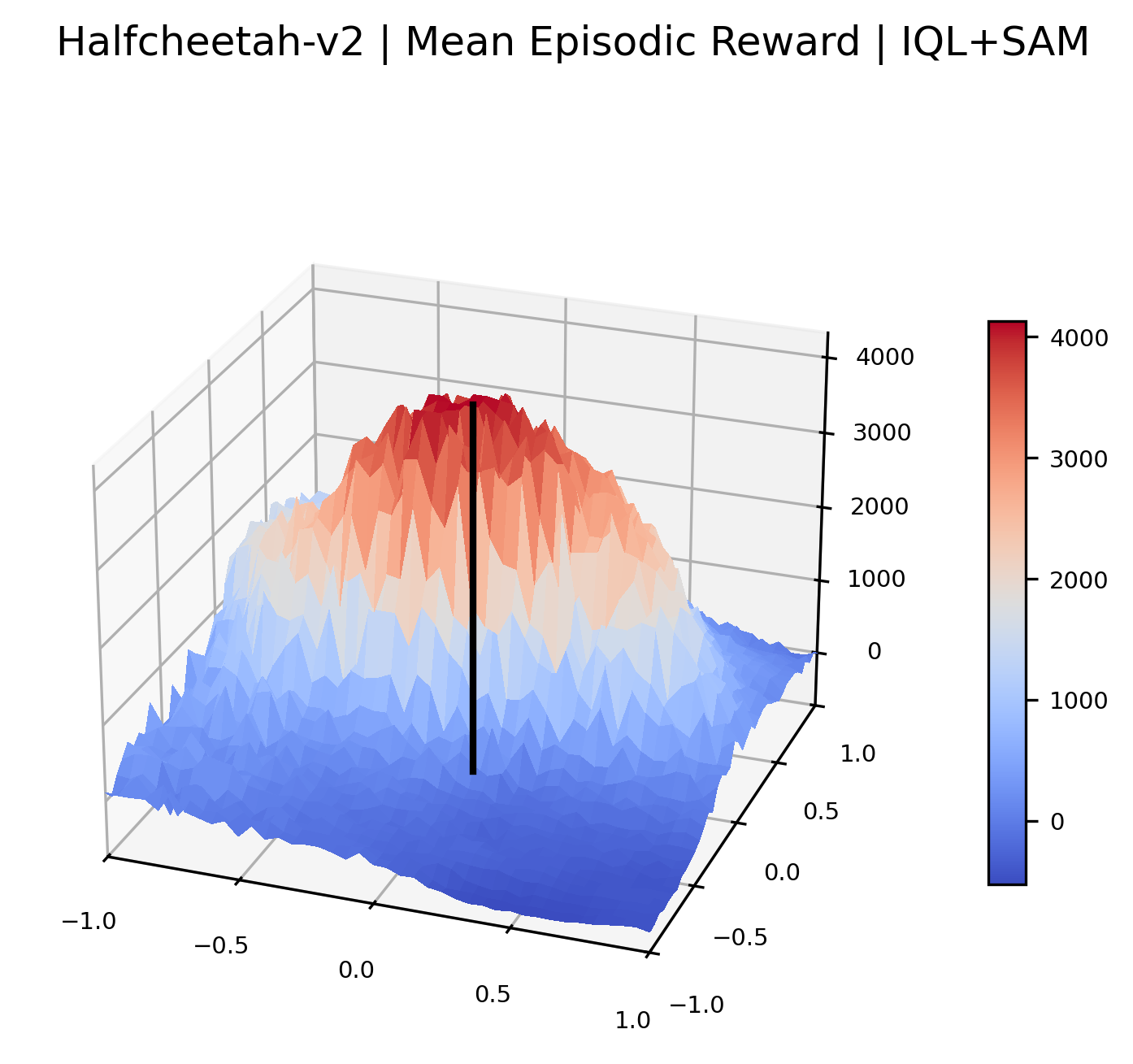}
    \caption{Reward surface visualization for IQL (Left) and IQL+SAM (Right) on HalfCheetah with random observation corruption. SAM produces a smoother landscape.}
    \label{fig:reward_surface}
\end{figure}

\paragraph{Reward Surface Visualization.}
To provide intuition, we visualize the reward surface learned by the agent, following the methodology of \cite{sullivan2022}. Figure \ref{fig:reward_surface} shows surfaces for IQL and IQL+SAM models trained on `halfcheetah` under random observation corruption. The standard IQL model converges to a region with sharp peaks and valleys, making it sensitive to perturbations. In contrast, IQL with SAM learns a significantly smoother and flatter reward surface. This visually confirms that SAM guides the agent to a more robust solution.

\begin{table}[ht]
\centering
\resizebox{\columnwidth}{!}{%
\begin{tabular}{@{}ll cc cc@{}}
\toprule
\multirow{2}{*}{Environment} & \multirow{2}{*}{Corruption} & \multicolumn{2}{c}{IQL} & \multicolumn{2}{c}{RIQL} \\
\cmidrule(lr){3-4} \cmidrule(lr){5-6}
& & Naïve & SAM (Ours) & Naïve & SAM (Ours) \\
\midrule
Halfcheetah & Observation & 30.32(4.29) & \textbf{32.42(2.26)} & \textbf{38.46(0.71)} & 35.92(3.10) \\
            & Mixture    & 12.68(0.48) & \textbf{16.72(1.28)} & \textbf{19.65(0.22)} & 18.19(3.50) \\
\midrule
Walker2d    & Observation & 21.19(12.58) & \textbf{31.86(3.96)} & 41.93(8.43) & \textbf{50.07(10.76)} \\
            & Mixture    & 12.31(1.25) & \textbf{12.61(4.27)} & 19.61(3.33) & \textbf{21.22(8.85)} \\
\midrule
Hopper      & Observation & 41.88(1.29) & \textbf{67.22(10.41)} & 49.12(2.99) & \textbf{65.34(15.09)} \\
            & Mixture    & 16.33(3.40) & \textbf{55.34(27.87)} & \textbf{60.40(11.67)} & 49.81(15.96) \\
\midrule
\multicolumn{2}{l}{Average score $\uparrow$} & 22.45 & \textbf{36.03} & 38.20 & \textbf{40.09} \\
\bottomrule
\end{tabular}%
}
\caption{Average Performance under adversarial corruption}
\label{tab:adversarial_corruption_styled}
\end{table}

\section{Conclusion}
We address the vulnerability of offline RL to data corruption by introducing Sharpness-Aware Minimization (SAM) as a plug-and-play optimizer. Applying SAM to strong baselines (IQL, RIQL) consistently and significantly improves performance on D4RL benchmarks under both random and adversarial corruption. Visualizations confirm SAM finds smoother reward surfaces, validating that optimizing for flatter minima is a promising direction for creating robust offline RL agents.

\section*{Acknowledgments}

We would like to extend our special thanks to Professor Jeff Schneider from Carnegie Mellon University for generously providing the essential computational resources for this research.

\bibliography{main} 

\appendix
\onecolumn
\section{Related Work}
In this section, we provide a more detailed discussion of the literature related to our work, covering robustness in offline reinforcement learning against data corruption and the principles of Sharpness-Aware Minimization.

\subsection{Robustness in Offline RL against Data Corruption}
The challenge of learning from corrupted offline datasets has been addressed through several distinct paradigms.

\paragraph{Heuristic-based Robust Algorithms.}
A primary line of research focuses on designing algorithms with specific mechanisms to mitigate the effects of corruption. The RIQL paper \cite{riql24} provides a foundational analysis, demonstrating that while many state-of-the-art offline RL algorithms fail, the learning scheme of Implicit Q-Learning (IQL) \cite{iql} exhibits a degree of inherent robustness. RIQL builds upon this by incorporating Huber loss and quantile estimators to specifically handle the heavy-tailed target distributions caused by dynamics corruption. However, as noted in our main paper, both IQL and RIQL still struggle with observation and mixture corruptions. This specific vulnerability serves as the primary motivation for our work.

\paragraph{Uncertainty and Consistency-based Methods.}
Other approaches aim to explicitly identify and down-weight corrupted data. For example, Uncertainty-Weighted MSG (UWMSG) \cite{uwmsg2023} leverages model ensemble uncertainty to assign lower weights to potentially noisy transitions. Similarly, TRACER \cite{tracer2023} learns a dynamics model and identifies corrupted data by detecting inconsistencies between the model's predictions and the actual transitions recorded in the dataset. These methods rely on the assumption that corrupted data will manifest as high uncertainty or inconsistency, which may not always hold true.

\paragraph{Sequence Modeling Approaches.}
With the rise of Transformer-based models, another direction has been to adapt sequence modeling for robust decision-making. Robust Decision Transformer (RDT) \cite{rdt2023} extends the Decision Transformer framework by incorporating randomization and consistency regularization to better handle corrupted trajectory data. This represents a different architectural approach to the problem.

\paragraph{Data Recovery Methods.}
A fundamentally different paradigm is to "clean" the dataset before the RL training process begins. For instance, ADG \cite{adg2025} proposes using diffusion models to recover the underlying clean data distribution from a corrupted offline dataset. While promising, this approach adds a significant pre-processing step and computational overhead.

\paragraph{Our Positioning.}
Our work distinguishes itself from the aforementioned literature by shifting the focus from data-level heuristics or filtering to the optimization process itself. Instead of designing mechanisms to handle specific corruption types (like RIQL) or trying to identify and remove bad data (like UWMSG and TRACER), we accept the corrupted dataset as given. We hypothesize that the core issue is that this corrupted data leads to a sharp and unstable loss landscape. Therefore, we propose modifying the optimizer with Sharpness-Aware Minimization to find a solution that is inherently robust to the noise induced by this data. This makes our method a general, plug-and-play module that addresses the geometric properties of the loss landscape, positioning it as a complementary approach to existing methods.

\subsection{Sharpness-Aware Minimization (SAM)}
Sharpness-Aware Minimization (SAM), introduced by \citet{pf2020}, is a powerful optimization technique designed to improve model generalization by finding parameters that lie in regions of the loss landscape with uniformly low loss values, i.e., "flat" minima. The core intuition is that models converging to sharp minima are less robust and generalize poorly, as small changes in the input data can lead to large changes in the output. SAM formalizes this by solving a minimax optimization problem:
\begin{equation}
    \min_{\theta} \max_{\|\epsilon\|_p \le \rho} L(\theta + \epsilon)
\end{equation}
where $L(\theta)$ is the loss function, and the inner maximization seeks a "worst-case" perturbation $\epsilon$ within a neighborhood of size $\rho$ that maximizes the loss. By minimizing this perturbed loss, the optimizer is explicitly guided away from sharp regions.

Subsequent research has further explored SAM's properties and proposed enhancements. \citet{wen2023doessharpnessawareminimizationminimize} provided a rigorous analysis of the exact sharpness notion that SAM regularizes, clarifying its underlying mechanism. To address the computational overhead of SAM, which requires two forward-backward passes per update, more efficient variants have been developed. For example, Sharpness-Aware Training for Free (SAF) \cite{du2023sharpnessawaretrainingfree} proposes a novel trajectory loss to mitigate sharp landscapes with almost no additional computational cost. These works have established SAM and its variants as effective, general-purpose tools for improving generalization across various domains in supervised learning. Our research is the first to systematically investigate and apply the SAM optimization framework to the problem of data corruption in the offline RL setting, demonstrating its effectiveness as a plug-and-play module to enhance agent robustness.

\section{Experimental Setup and Implementation Details}

\subsection{Data Corruption Details}
To rigorously evaluate the robustness of our proposed method, we employ two distinct modes of data corruption on the D4RL datasets: random and adversarial. For any given experiment, a fixed portion of the dataset (30\% in our main experiments) is selected for corruption.

\paragraph{Random Corruption.}
For a randomly selected subset of transitions $(s, a, r, s')$, we apply perturbations as follows:
\begin{itemize}
    \item \textbf{Observation Corruption:} The state $s$ is perturbed to $\hat{s} = s + \lambda \cdot \text{std}(S)$, where $\lambda$ is a vector sampled uniformly from $[-\epsilon, \epsilon]^{d_s}$ and $\text{std}(S)$ is the element-wise standard deviation of all states in the dataset.
    \item \textbf{Mixture Corruption:} In this setting, we sequentially apply observation, action, and reward corruptions to independently chosen subsets of the data.
\end{itemize}

\paragraph{Adversarial Corruption.}
For adversarial attacks, we first pre-train a baseline agent to obtain a set of Q-functions, $Q_p$. We then craft perturbations to maximize the damage to the agent's performance.
\begin{itemize}
    \item \textbf{Observation Corruption:} For a selected state $s$, the perturbation is found by solving $\hat{s} = \min_{\hat{s} \in \mathbb{B}(s, \epsilon)} Q_p(\hat{s}, a)$, where $\mathbb{B}(s, \epsilon)$ is a ball of radius $\epsilon \cdot \text{std}(S)$ around $s$.
    \item \textbf{Mixture Corruption:} In this setting, we sequentially apply observation, action, and reward corruptions to independently chosen subsets of the data.
\end{itemize}
The optimization problems for generating adversarial states and actions are solved using Projected Gradient Descent (PGD) \cite{madry2019deeplearningmodelsresistant}.

\subsection{Algorithm Pseudocode}
Our approach involves integrating the SAM optimizer into the training loop of IQL-family algorithms. As determined by our ablation studies (see Appendix C), we apply SAM exclusively to the update of the value function network (Vf). Algorithm \ref{alg:sam_iql} outlines this process.

\begin{algorithm}[h]
\caption{IQL/RIQL Training with SAM Optimizer for Vf}
\label{alg:sam_iql}
\begin{algorithmic}[1]
\STATE Initialize actor $\pi_\phi$, Q-functions $Q_{\theta_i}$, value function $V_\psi$.
\STATE Initialize standard Adam optimizers for $\pi_\phi$ and $Q_{\theta_i}$.
\STATE Initialize SAM optimizer for $V_\psi$, wrapping an Adam optimizer.
\FOR{each training step}
    \STATE Sample a minibatch $\mathcal{B} = \{(s, a, r, s', d)\}$ from dataset $\mathcal{D}$. \textit{// Q-Function and Actor update (standard)}
    
    \STATE Compute Q-function loss $L_Q$ on $\mathcal{B}$.
    \STATE Update $Q_{\theta_i}$ with Adam optimizer.
    \STATE Compute actor loss $L_\pi$ on $\mathcal{B}$.
    \STATE Update $\pi_\phi$ with Adam optimizer. \textit{// Value Function update (with SAM)}
    
    \STATE Compute value function loss $L_V$ on $\mathcal{B}$.
    \STATE $L_V$.backward()
    \STATE \textbf{v\_optimizer.first\_step(zero\_grad=True)} \COMMENT{SAM: Ascend step}
    \STATE Compute value function loss $L_V^{adv}$ on $\mathcal{B}$ using perturbed weights.
    \STATE $L_V^{adv}$.backward()
    \STATE \textbf{v\_optimizer.second\_step(zero\_grad=True)} \COMMENT{SAM: Descend step}
    
    \STATE Update target networks.
\ENDFOR
\end{algorithmic}
\end{algorithm}

\subsection{Hyperparameter Settings}
Our implementation builds upon the publicly available source code of RIQL \cite{riql24}. To ensure a fair comparison, all hyperparameters for the base IQL and RIQL algorithms (e.g., learning rates, network architectures, discount factor) are kept identical to those in the original RIQL codebase. The only modification is the replacement of the Adam optimizer for the value function with our SAM optimizer.

The key hyperparameter for SAM is the neighborhood radius, $\rho$. The specific values of $\rho$ used for the main results presented in Table 1 and Table 2 of the main paper are detailed in Table \ref{tab:rho_values} below. Notably, for a given environment, the same $\rho$ value was used for both Observation and Mixture corruption types, indicating that the choice of $\rho$ is primarily environment-dependent.

\begin{table}[h]
\centering
\caption{Selected $\rho$ values for SAM in main experiments.}
\label{tab:rho_values}
\begin{tabular}{llcc}
\toprule
\textbf{Algorithm} & \textbf{Environment} & \textbf{$\rho$ (Random Corr.)} & \textbf{$\rho$ (Adversarial Corr.)} \\
\midrule
IQL+SAM & HalfCheetah & 1.0 & 1.0 \\
IQL+SAM & Walker2d & 0.5 & 0.1 \\
IQL+SAM & Hopper & 0.1 & 0.5 \\
\midrule
RIQL+SAM & HalfCheetah & 1.0 & 0.1 \\
RIQL+SAM & Walker2d & 0.1 & 0.1 \\
RIQL+SAM & Hopper & 1.0 & 1.0 \\
\bottomrule
\end{tabular}
\end{table}

\section{Ablation Studies and Additional Results}

\subsection{Ablation Study: SAM Component and Hyperparameter $\rho$}
To determine the most effective way to integrate SAM and to understand the sensitivity of its main hyperparameter $\rho$, we conducted an extensive ablation study. Experiments were performed on the `halfcheetah-medium-replay-v2` dataset with random observation corruption. We tested applying SAM to the Actor (A), Q-function (Q), and V-function (V) networks, both individually and in combination. The results are summarized in Table \ref{tab:sam_ablation_booktabs}.

Our findings clearly indicate that the greatest performance gains are achieved when SAM is applied to the V-function. Applying SAM solely to the V-function (column 'V') yields the highest average score (29.03), a notable improvement over the baseline RIQL (26.03). Combinations including the V-function ('AV' and 'QV') also show performance gains, whereas applying SAM to the Actor or Q-function alone ('A', 'Q') or together ('AQ') results in a performance drop. This provides strong empirical evidence for our design choice in the main paper to exclusively apply SAM to the V-function's optimization process.

Furthermore, the results demonstrate the importance of tuning the neighborhood radius $\rho$. For the best-performing 'V' configuration, performance peaks at $\rho=0.15$ (30.71), highlighting that the effectiveness of SAM is sensitive to this hyperparameter.

\begin{table}[ht]
\centering
\caption{Ablation study on which component to apply SAM and the effect of hyperparameter $\rho$. All experiments are on `halfcheetah-medium-replay-v2` with random observation corruption. Results are reported as mean(std) normalized scores over 3 seeds. A, Q, V denote Actor, Q-function, and V-function respectively.}
\label{tab:sam_ablation_booktabs}
{\small
\begin{tabular}{@{}cc ccccccc@{}}
\toprule
\multirow{2}{*}{rho} & \multirow{2}{*}{RIQL} & \multicolumn{7}{c}{RIQL+SAM} \\ \cmidrule(lr){3-9} 
                     &                       & A           & Q           & V           & AQ          & AV          & QV          & AQV         \\ \midrule
0.05                 & \multirow{4}{*}{26.03(3.61)} & 24.29(7.52) & 19.90(4.30) & 26.29(4.36) & 21.76(2.97) & 27.16(4.30) & 25.81(1.94) & 23.96(1.63) \\
0.1                  &                       & 24.32(2.39) & 23.76(3.05) & 29.35(1.36) & 21.33(1.80) & 29.62(2.17) & 29.74(3.76) & 27.70(4.72) \\
0.15                 &                       & 22.54(2.37) & 23.03(2.00) & 30.71(2.56) & 19.28(0.57) & 31.13(1.41) & 26.90(2.19) & 26.71(2.21) \\
0.2                  &                       & 20.58(1.21) & 25.23(2.49) & 29.76(0.86) & 23.39(3.78) & 22.57(2.75) & 27.80(5.10) & 26.33(3.55) \\ \midrule
Average Score        & 26.03                 & 22.93       & 22.98       & 29.03       & 21.44       & 27.62       & 27.56       & 26.18       \\
Improvement          & 0                     & -3.1        & -3.05       & 3           & -4.59       & 1.59        & 1.53        & 0.15        \\ \bottomrule
\end{tabular}
}
\end{table}

\subsection{Ablation Study: Batch Size}
SAM literature suggests a connection between batch size and the sharpness of the loss landscape, with smaller batches sometimes being more effective at navigating to flatter minima. We investigate this effect in the context of offline RL. Table \ref{tab:batch_size_ablation} shows the performance of RIQL+SAM (applied to Vf) under different batch sizes. The experiment was conducted on `halfcheetah-medium-replay-v2` with random observation corruption and a fixed $\rho=1$.

The results indicate a non-trivial relationship. While very small batch sizes (e.g., 1-8) lead to unstable training and poor performance, a batch size of 64 yields the best average score (34.17), outperforming the standard batch size of 256 (33.74). This supports the hypothesis that a moderately reduced batch size can be beneficial for SAM, likely by introducing noise that helps escape sharper local minima. However, there is a clear trade-off, as excessively small batches destabilize the learning process.

\begin{table}[ht]
\centering
\caption{Ablation Study on Batch Size for RIQL+SAM (Vf only, $\rho=0.15$) on `halfcheetah-medium-replay-v2` with random observation corruption.}
\label{tab:batch_size_ablation}
\begin{tabular}{@{}lcc@{}}
\toprule
Batch\_size   & Norm\_score(avg)      & Norm\_score(org)      \\ \midrule
256           & 33.74(2.48)           & \textbf{35.30(3.44)} \\
128           & 28.48(6.21)           & 26.45(5.60)          \\
64            & \textbf{34.17(1.42)} & 33.42(3.77)          \\
32            & 31.05(1.28)           & 32.50(2.58)          \\
16            & 33.69(3.83)           & 34.43(4.95)          \\
8             & 24.49(14.15)          & 24.85(13.46)         \\
4             & 22.89(13.85)          & 20.16(14.19)         \\
2             & 9.66(14.51)           & 8.46(12.45)          \\
1             & 21.57(10.83)          & 23.55(17.64)         \\ \midrule
RIQL(baseline) & 26.03(3.61)           & 25.52(6.30)          \\ \bottomrule
\end{tabular}
\end{table}

\subsection{Additional Results: Varying Corruption Levels}
To verify that the performance benefits of SAM are not confined to a single corruption setting, we conducted additional experiments on the `halfcheetah-medium-replay-v2` dataset by varying both the corruption rate and range. The results, presented in Table \ref{tab:corruption_rate} and Table \ref{tab:corruption_range}, demonstrate that IQL+SAM consistently outperforms the IQL baseline across all tested levels of random observation corruption. This indicates that SAM provides a robust advantage, enhancing performance across a spectrum of data quality, rather than being tuned for a specific corruption intensity.

\begin{table}[ht]
\centering
\caption{Performance on different corruption rates ($\epsilon=1.0$).}
\label{tab:corruption_rate}
\begin{tabular}{@{}lllcc@{}}
\toprule
Environment & Attack Element & Corruption Rate & IQL           & IQL+SAM              \\ \midrule
\multirow{4}{*}{Halfcheetah} & \multirow{4}{*}{Observation} & 0.2 & 28.23(1.32) & \textbf{32.23(4.20)} \\
                             &                              & 0.3 & 21.01(3.01) & \textbf{30.29(0.63)} \\
                             &                              & 0.4 & 18.15(1.49) & \textbf{23.73(1.02)} \\
                             &                              & 0.5 & 15.23(1.67) & \textbf{19.98(3.54)} \\ \midrule
\multicolumn{3}{l}{Average Score}                         & 20.66         & \textbf{26.56}       \\ \bottomrule
\end{tabular}
\end{table}

\begin{table}[ht]
\centering
\caption{Performance on different corruption ranges (rate=0.3).}
\label{tab:corruption_range}
\begin{tabular}{@{}lllcc@{}}
\toprule
Environment & Attack Element & Corruption Range & IQL           & IQL+SAM              \\ \midrule
\multirow{4}{*}{Halfcheetah} & \multirow{4}{*}{Observation} & 0.25 & 35.85(2.57) & \textbf{38.51(0.90)} \\
                             &                              & 0.5  & 31.76(1.33) & \textbf{33.62(1.84)} \\
                             &                              & 1    & 21.01(3.01) & \textbf{30.29(0.63)} \\
                             &                              & 2    & 17.57(1.41) & \textbf{22.28(1.24)} \\ \midrule
\multicolumn{3}{l}{Average Score}                          & 26.55         & \textbf{31.18}       \\ \bottomrule
\end{tabular}
\end{table}

\end{document}